\documentclass{article}

\usepackage{arxiv}

\usepackage[utf8]{inputenc} 
\usepackage[T1]{fontenc}    
\usepackage{hyperref}       
\usepackage{url}            
\usepackage{booktabs}       
\usepackage{amsfonts}       
\usepackage{nicefrac}       
\usepackage{microtype}      
\usepackage{lipsum}

\usepackage{mathtools}
\usepackage{caption}
\usepackage{subcaption}
\usepackage{amsmath}
\usepackage{float}
\usepackage{graphicx}
\usepackage{hyperref}

\title{Impact of Batch Normalization on Convolutional Network Representations}

\author{
  Hermanus L. Potgieter $^{1,2,4}$\quad Coenraad Mouton $^{1,2}$\quad Marelie H. Davel $^{1,2,3}$ \\
  $^1$North-West University 
\quad $^2$CAIR
\quad $^3$NITheCS
\quad $^4$SANSA \\
\texttt{potgieterharmen@gmail.com}
\thanks{This work is a preprint of a published paper by the same name \cite{potgieter2024impact}. The authenticated version is available online at \url{https://doi.org/10.1007/978-3-031-78255-8_14}}}

\makeatletter
\def\thanks#1{\protected@xdef\@thanks{\@thanks
        \protect\footnotetext{#1}}}
\makeatother

\begin{document}
\maketitle

\begin{abstract} 
Batch normalization (BatchNorm) is a popular layer normalization technique used when training deep neural networks. 
It has been shown to enhance the training speed and accuracy of deep learning models. However, the mechanics by which BatchNorm achieves these benefits is an active area of research, and different perspectives have been proposed.
In this paper, we investigate the effect of BatchNorm on the resulting hidden representations, that is, the vectors of activation values formed as samples are processed at each hidden layer.
Specifically, we consider the sparsity of these representations, as well as their implicit clustering  -- the creation of groups of representations that are similar to some extent. 
We contrast image classification models trained with and without batch normalization and highlight consistent differences observed. 
These findings highlight that BatchNorm's effect on representational sparsity is {\it not} a significant factor affecting generalization, while the representations of models trained with BatchNorm tend to show more advantageous clustering characteristics.
\end{abstract}


\section{Introduction}
Deep learning has become a particularly important set of machine learning techniques and is widely applied to solve real-world tasks. 
At the same time, many open questions remain with regard to the ability of these deep neural networks (DNNs) to generalize so well, that is, their ability to perform well on unseen data.
Although there is not yet a theoretical framework to assist us in reasoning about these models~\cite{zhang2021understanding}, the generalization ability of DNNs has been studied from many perspectives, such as the geometry of the loss landscape \cite{Foret2020SharpnessAwareMF}, statistical measures of stability and robustness \cite{bousquet2000algorithmic}, size of margins (distance to the decision boundary between classes)~\cite{mouton2023input}, and information-theoretic techniques~\cite{kawaguchi2023does}, among others. 
  
A promising research direction is to study the characteristics of the internal data representations formed by DNNs, where
each representation is the vector of activation values from a specific layer for a given sample.
Aspects of these representations that have been studied include the size of margins in the representation space~\cite{optimal_transport,predict_gen_margin,natekar2020representation}; 
the `quality' of representations, evaluated using the consistency of class-specific representations and their robustness when combined~\cite{natekar2020representation}; 
and representation sparsity, that is, the number of non-zero elements in a data representation~\cite{glorot_2011}. 
In this work, we also study the characteristics of the internal representations of DNNs, but focus on the effect that a very specific technique -- Batch Normalization (BatchNorm) -- has on internal representation quality.

BatchNorm~\cite{batchnorm_original} is a popular technique used to normalize hidden activations when training DNNs. Networks trained with BatchNorm show desirable properties such as faster convergence and better generalization ability~\cite{liexponential_bn,santurkar_bn}. Despite the success and widespread adoption of BatchNorm, the exact mechanisms by which BatchNorm achieves its performance remain unclear. In this work, we study the effect of BatchNorm on the internal representations of convolutional neural networks (CNNs). More precisely, we analyze the resulting sparsity and clustering characteristics of CNN activations for models trained with and without BatchNorm.
%
We selected these two elements, specifically:
\begin{itemize}
    \item Representational sparsity is generally considered a desirable property for DNNs, and is considered to be one of the reasons for the performance of rectified linear unit (RELU) activation functions~\cite{glorot_2011}.
    Although the sparsity of neural networks parameters (parametric sparsity) is well studied, the sparsity of the internal representations (representational sparsity) is less explored~\cite[p.~251]{Goodfellow-et-al-2016}. 
    The effect of BatchNorm on representational sparsity has been studied to a limited extent for Multilayer Perceptrons (MLPs)~\cite{pretorius2019relu} but no such analysis has yet been performed for CNNs.

\item 
The clustering characteristics of representations have shown to be a promising avenue to better understand generalization in DNNs~\cite{optimal_transport,liao2016learning,natekar2020representation}.
As such, comparing the clusters formed by models trained with and without BatchNorm could provide insights into how BatchNorm increases generalization performance.
\end{itemize}
In summary, the objective of this paper is to compare models trained with and without BatchNorm and to analyze the effect of BatchNorm on the hidden representations. 


\section{Background}
This section introduces the BatchNorm technique as well as current views on the rationale for its performance. 
We also review current perspectives on the effect of representation sparsity and clustered representations on generalization performance.

\subsection{Batch Normalization}
\label{sec:Background_Batch_norm}
Batch normalization~\cite{batchnorm_original} is a popular layer normalization technique which speeds up training and improves the performance of deep neural networks. 
The goal of BatchNorm is to normalize each batch of activation values for some layer to a learned mean and standard deviation during both training and inference. 
More precisely, for the activation values $\mathbf{x}^l$ at layer $l$ for a sample $\mathbf{x}$, the BatchNorm operation is defined as:
\begin{equation}
\mathrm{BN}(\mathbf{x}^l)=\boldsymbol{\gamma} \odot \frac{\mathbf{x}^l-\hat{\boldsymbol{\mu}}^l_{\mathcal{B}}}{\hat{\boldsymbol{\sigma}}^l_{\mathcal{B}}}+\boldsymbol{\beta} .\end{equation}
where $\hat{\boldsymbol{\mu}}^l_{\mathcal{B}}$ and $\hat{\boldsymbol{\sigma}}^l_{\mathcal{B}}$ is the \textit{per feature} mean and standard deviation vectors calculated over a batch of samples $\mathbf{X}_\mathcal{B}$. The vectors $\boldsymbol{\gamma}$ and $\boldsymbol{\beta}$ are learnable parameters of the same dimensionality as $\mathbf{x}^l$. 
In the case of CNNs, each channel is considered a `feature', and the mean and standard deviation are calculated independently per channel. More precisely, the activation values for a convolution layer $l$ for a batch of samples can be denoted as a 4D tensor such that $\mathbf{X}_{\mathcal{B}}^l \in \mathbb{R}^{N\times C\times H\times W}$. Here $N, C, H$, and $W$ correspond to the number of samples in the batch, the number of channels, and the height and width of the representation, respectively. The per-channel mean and standard deviation are calculated in the same way but for a channel instead of a node.


Despite the widespread popularity of BatchNorm, there is not yet a consensus on the exact mechanism by which BatchNorm speeds up training and improves generalization. 
Originally the success of BatchNorm was motivated by its reduction of `internal covariance shift'~\cite{batchnorm_original}, meaning that BatchNorm reduces distributional differences between the inputs to different layers in the network. However, Santurkar et al.~\cite{santurkar_bn} show that even when the internal distributions are artificially shifted (by injecting noise after BatchNorm layers) there is no noticeable degradation in performance. Instead, they argue from an optimization perspective, and conclude that BatchNorm drastically smooths the loss landscape. From a similar perspective, Bjorck et al.~\cite{bjorck_understanding_2018} show that without BatchNorm, a small subset of the activations in deeper layers grow uncontrollably large during training. They further show that BatchNorm prevents this phenomenon and allows for the use of larger learning rates that result in speedier convergence. Similarly, Li and Arora~\cite{liexponential_bn} show that an exponential learning rate schedule is made possible by BatchNorm. 

Although the optimization perspective is the dominant view, several other authors have also investigated the benefits of BatchNorm from alternative perspectives.  These include the effect of BatchNorm on adversarial robustness~\cite{benz_bn}, its effect on the density of linear regions in the input space~\cite{balestriero_bn}, and the orthogonality of internal representations~\cite{daneshmand_bn}. 


\subsection{Sparsity}
Representational sparsity refers to the extent to which a representation consists of values that are zero or close to zero~\cite[p.~251]{Goodfellow-et-al-2016}.
Glorot et al.~\cite{glorot_2011} approach sparsity from a biological perspective, arguing the use of rectified neurons over sigmoidal neurons is inspired by neuroscience. In addition to the use of rectified neurons, they propose the use of $L_1$ regularization on the activation values to promote sparsity as a mechanism for better performance. 
Sparsity has been attributed with benefits such as information disentangling, 
efficient variable-size representation, and
representations being more linearly separable~\cite{glorot_2011}.
Representational sparsity can thus be seen as desirable in CNNs
~\cite{Bengio_sparsity}.

Pretorius et al.~\cite{pretorius2019relu} investigated the effect of activation functions on Multi-Layer Perceptrons (MLPs) by analysing network characteristics such as network sparsity.
They found that networks trained with ReLU activation are more sparse on average than networks trained on sigmoidal activations.
They also investigated the effect of BatchNorm on representation sparsity and found that the models trained with BatchNorm tend to have {\em less} sparse representations. 
It was also found that the less sparse BatchNorm models tend to have better generalization performance, and it was suggested that
sparsity is not necessarily an indication of good generalization.

\subsection{Clustered Representations}

Clustering is an unsupervised learning method that aims to form groups of similar elements based on different metrics.
Methods can range from forming clusters by minimizing or maximizing a quality measure~\cite{macqueen1967some}, using hierarchical methods that build a dendrogram to iteratively merge or split clusters based on a distance metric~\cite{nielsen_hierarchical_2016}, or using linear algebra directly (eigenvalues and eigenvectors) to cluster~\cite{shi_normalized_2000}.

\subsubsection{Class-based Clustering}
A neural network can be conceptualized as a clustering algorithm, as it inherently groups data with similar features together within its hidden layers~\cite{Mouton_layerwise}, meaning similar samples are modelled in a similar way.
Since the network is trained to predict the correct class of a given sample, it tends to cluster samples from the same class together. 
That said, the initial layers often capture features that are shared across different classes, leading to inter-class clustering. 
However, as observed by Caron et al. \cite{caron2018deep}, deeper layers typically contain more class-specific information, resulting in distinct clusters that align closely with the true class labels.
Class-based cluster analysis therefore assigns class labels as cluster identifiers and evaluates cluster characteristics.

\subsubsection{Class-Agnostic Clustering}
An alternative viewpoint is not to consider the class label information at all but rather to consider the purity of clusters formed at different layers of a network. In this case, the number of clusters that are relevant is not known ahead of time.

\subsubsection{Clustering Techniques}
One of the most popular clustering methods is k-means clustering~\cite{macqueen1967some}. 
The method partitions the samples into $k$ clusters. 
The distance to each cluster centre is calculated, and the sample is placed in the cluster with the nearest mean.
After each iteration, the cluster centres are moved to the centre of the points assigned to the cluster. 
The process is iterated for a number of steps. 
Different variations of the method exist such as k-means++~\cite{kmeans++}, Fuzzy C-means clustering \cite{dunn1973fuzzy}, and k-medians clustering \cite{jain1988algorithms}, all of which use a similar principle as k-means clustering. 
The k-means++ algorithm improves upon k-means with a more intelligent selection of the initial centroids. 
Fuzzy C-means allow data points to belong to multiple clusters, assigned via the similarity to the centroid. 
It focuses more on the membership relationship and uncertainty of the data points. 
K-medians minimise the sum of absolute differences between the centroids and data points, which is a more robust approach.
The number of clusters must be specified in advance for the k-means method, which often necessitates a hyperparameter search to determine the optimal number of clusters.

\subsubsection{Evaluating cluster quality}
\label{sec:dbi}
The Davies-Bouldin index (DBI)~\cite{davies1979cluster} is a popular metric to evaluate cluster quality.
It evaluates the average similarity between each cluster and its most similar cluster, offering an indication of how well-separated and cohesive the clusters are. 
It is computed by comparing the average distance between clusters and the size of the clusters. 
This method can be used to compare the relative appropriateness of different data partitions. 
Importantly, the index is not sensitive to the number of clusters or the specific partitioning of the data.
The DBI is defined as: 
\begin{equation}
    DBI  = \frac{1}{k}\sum_{i = 1}^{k}\max_{i\neq j}\left\{\frac{d(x_{i}) + d(x_j)}{d(c_i,c_j)}\right\}
\end{equation}
where $k$ denotes the number of clusters, $i, j$ are cluster labels, $d(x_{i})$ and $d(x_j)$ are the mean intra-cluster distance for the clusters $i$ and $j$, respectively, and $d(c_i,c_j)$ is the distance between the cluster centroids.
The DBI can be used to measure the purity of the of representational clusters based on the class that the representation belongs to.

\subsubsection{Clustered representations and generalization}

There are limited studies that explore the link between the characteristics of representation clusters and generalization directly.
The two main papers that make a direct link between these concepts are those by Natekar and Sharma~\cite{natekar2020representation} and Liao et al.~\cite{liao2016learning}. 

Natekar and Sharma proposed three measures of quality of representations to predict generalization in a post-hoc diagnostic setting \cite{natekar2020representation}. The first metric they measure is consistency, which is particularly relevant, as it is defined as an indication of representation cluster purity. 
To measure consistency, the DBI is calculated on a dimensionally reduced version of the activation values of a layer. This produces a measure of how consistent the representations are within a cluster, as well as how different representations are across clusters. 
Dimensionality reduction is performed with principal component analysis (PCA) and max pooling. 
The main observation from their study was that simple measures based on the quality of internal representations are somewhat predictive of the generalization performance and they, therefore, conjecture that clustered representations of objects in internal layers aid in generalization.

Liao et al. \cite{liao2016learning} use the process of clustering representations as a regulariser, and demonstrate some performance gains in the process. 
That is, they adjust the loss function during optinization to encourage cluster-forming.
They looked at three different clustering methods namely: sample clustering, spatial (channel) clustering and co-clustering of both samples and channels. 
The regularisation term that they add to the standard loss consists of the distance of the element to the cluster mean. 
The cluster centres are initialised and updated with k-means, and then resampled and updated through the $k$-$means++$ procedure. 
Their results across clustering techniques were fairly similar, with sample and spatial clustering outperforming co-clustering with a small margin.

Clustering has been studied more extensively in the unsupervised learning setting.
Caron et al.~\cite{caron2018deep} use the clustering of internal representations to train CNNs. 
The training process consists of an alternating process of clustering internal representations (using standard k-means) and updating the parameters until convergence. 
Once the clustering is complete, the cluster tags are used as pseudo-labels for the corresponding hidden representation vectors. 
The pseudo-labels are used to train a classifier network on top of the CNN to predict the pseudo-labels for each hidden representation vector. 
This idea was further developed by Zhan et al.~\cite{zhan2020online} who introduced a batch-aware version that addresses instability caused by samples changing their batch membership; 
and by Li et al.~\cite{li2020prototypical} who forced the training process to create representations that are close to 
the cluster centroids.

\section{Experimental setup}

In this section, we describe the approach used to investigate the effect of BatchNorm on CNN representations. 
We select a set of 
CNN architectures and datasets and develop comparative models. We then develop metrics of sparsity and techniques to evaluate the cluster characteristics of representations.


\subsection{Models}
\label{sec:exp_models}

We consider two different datasets and network architectures for experimentation. 
For datasets, we select MNIST~\cite{lecun1998mnist} and CIFAR10~\cite{krizhevsky2009learning}. MNIST, while simple, is useful for probing generalization from a clustering perspective as it consists of well-separated classes. On the other hand, CIFAR10 consists of more natural images and a greater amount of overlap among classes. This allows us to verify whether results on MNIST also hold on CIFAR10. 
We select architectures that match the datasets: a simple CNN with $4$ convolutional layers trained on the MNIST dataset, and a standard, deeper architecture with $13$ convolutional layers trained on the CIFAR10 dataset. 
For each of these architectures, we train a model with and without BatchNorm using $4$ different random initialization seeds each. 

\subsubsection{MNIST}

For the MNIST dataset, we make use of an architecture similar to the `standard architecture' described by Nakkiran et al.~\cite{nakkiran_deep_2019}.  
This $5$-layer CNN consists of $4$ convolutional layers with [$16, 32, 64, 128$] channels, respectively. There are also max pooling layers after the $2$nd and $3$rd convolutional layers with a stride of $2$. We alter the architecutre by adding a fully connected hidden layer of $100$ nodes preceding the output layer of $10$ nodes, as we are interested in the representations formed by both the convolutional and fully connected layers. 
This is in contrast to the architecture of Nakkiran et al.~\cite{nakkiran_deep_2019} which only has a single fully connected output layer of $10$ nodes.

Hyperparameters are chosen based on a hyperparameter sweep that tracks 
validation accuracy at the point of interpolation.
We apply a ReLU activation after each hidden layer, cross-entropy as loss function, and SGD as optimizer for both the BatchNorm and non-BatchNorm models. We use a batch size of $64$, a learning rate of $0.01$, a learning rate schedule that multiplies the learning rate with $0.99$ every $10$ epochs, and train for $1\ 000$ epochs.
We also include a Nesterov momentum of $0.99$. 
We terminate the optimization once the network interpolates the training data (reaches $100\%$ training accuracy).
The accuracy achieved during training is displayed in Table \ref{tab:Training_accuracy}. Both sets of models interpolated on the training data with the BatchNorm models performing a bit better on the validation and test sets than the non-BatchNorm. The variation between seeds is minimal for both the BatchNorm and non-BatchNorm models.
\subsubsection{CIFAR10}

For the CIFAR10 dataset, we rely on the well-known VGG-16~\cite{simonyan2014very} architecture. This architectures consists of $13$  ReLU-activated convolutional layers with $[64, 64, 128, 128, 256, 256, 256, $
$512, 512, 512, 512, 512, 512]$ channels and a kernel size of $3\times 3$. There are also $2\times 2$ max pooling layers after the $2^{nd}$, $4^{th}$, $7^{th}$, $10^{th}$, and $13^{th}$ convolutional layers with a stride of $2$. 
We 
do not use the hyperparameters from the original implementation of Simonyan and Zisserman~\cite{simonyan2014very}
but rather,  
perform a hyperparameter sweep using the validation set.
This is done to achieve as close to current benchmarks on the dataset and architecture as possible.
Specifically, we make use of cross-entropy loss in combination with the ADAM optimizer. We use an initial learning rate of $0.001$ for the models with BatchNorm, and a smaller learning rate of $0.0001$ for those without. The learning rate is multiplied by $0.99$ after each epoch, and the momentum and stability terms of ADAM are kept at the default values. We use the same batch size of $64$ for both sets of models. 
We train for $300$ epochs using early stopping with a patience of $20\%$ on the validation error.

The accuracy for both sets of models is in Table \ref{tab:Training_accuracy}. Both sets of models achieved near interpolation on the training data, with the BatchNorm models significantly outperforming the non-BatchNorm models on the unseen data from both the training and evaluation sets. Additionally, the BatchNorm models exhibited minimal variation between seeds, whereas the accuracy of the non-BatchNorm models showed much greater variability across different seeds on the unseen data.


\begin{table}[H]
\centering
\caption{Train, validation, and evaluation classification accuracy for the BatchNorm (BN) and non-BatchNorm (NBN) models on MNIST and CIFAR10. $\pm$ indicates the standard deviation across $4$ random initialization seeds.}
\label{tab:Training_accuracy}
\begin{tabular}{c|ccc}
\multicolumn{1}{l}{} & \multicolumn{1}{l}{\textbf{Training Acc}} & \multicolumn{1}{l}{\textbf{Validation Acc}} & \multicolumn{1}{l}{\textbf{Evaluation Acc}}    \\ 
\hline
CNN MNIST BN         & $100.000\pm{0.000}$                               & $99.355\pm{0.010}$                  & $99.198\pm{0.100}$                            \\
CNN~ MNIST NBN       & $100.000\pm{0.000}$                               & $99.250\pm{0.087}$                  & $99.150\pm{0.072}$                            \\
VGG-16 CIFAR10 BN     & $99.988\pm{0.024}$                        & $87.080\pm{0.315}$                          & $85.99\pm{0.320}$                             \\
VGG-16 CIFAR10 NBN    & $99.982\pm{0.021}$                        & $81.030\pm{1.008}$                          & $79.435\pm{1.059}$                              
\end{tabular}
\end{table}

\subsection{Sparsity}
\label{sec:Exp_sparsity}

We only use the training data representations during analysis, as we are specifically focused on data that both models have seen and learned. 
This approach provides an indication of where these two models differ, given that it is data both have memorized.
To extract the representations of the samples, we pass each sample through the network and record the activation values of each layer after the ReLU activation function.
This produces a representation  $\mathbf{x}^l \in [0, \infty]$ for each sample at each layer of the network, and results in the 4D tensor $\mathbf{X}_{\mathcal{B}}^l$ for a batch of samples, as described in Section~\ref{sec:Background_Batch_norm}.
When analysing a certain dimension, such as the layer or channels of the network,  the tensor is unfolded into a 2D matrix where the one dimension is the size of the element analysed and the other is the product of all other dimensions. For example, analysing the sample dimension gives a matrix of $\mathbf{X}_{\mathcal{B}}^l \in \mathbb{R}^{N\times CHW}$.

We define a sparse element as any element with a value of zero.
While activation values close to zero could have minimal impact on the network's decision-making process and could therefore also be considered as values for which the network effectively does not activate~\cite[p.~251]{Goodfellow-et-al-2016}, we 
do not consider such small values as sparse elements.
To calculate sparsity, we first identify the inactive elements -- those with activation values of zero.
The tensor is then reshaped to focus on the specified axis. 
For channel sparsity, the resulting 2D matrix is $\mathbf{X}_{\mathcal{B}}^l \in \mathbb{R}^{C \times NHW}$ and the sparsity calculated as:
\begin{equation}
\label{eq:Sparsity_avg_channel}
    s_c = \frac{1}{N \cdot H \cdot W} \sum_{j=1}^{NHW} \mathbb{I}(\mathbf{x}^{l}_{c,j} = 0)
\end{equation}
where $\mathbf{x}^{l}_{c,j}$ is the element of the matrix at position $(c, j)$ where $c$ is the channel, and $\mathbb{I}(\cdot)$ is the indicator function, which equals 1 if the condition inside is true and 0 otherwise.
To calculate the layer sparsity we use the vector $\mathbf{x}_{\mathcal{B}}^l \in \mathbb{R}^{1 \times NCHW}$ as follows:
\begin{equation}
\label{eq:Sparsity_avg_layer}
    s_l = \frac{1}{N \cdot C \cdot H \cdot W} \sum_{j=1}^{NCHW} \mathbb{I}(\mathbf{x}^{l}_{j} = 0)
\end{equation}
resulting in a single sparsity value per layer (per model and seed).

\subsection{Clustered Representations}
\label{sec:Exp_cluster}
The same representations ($\mathbf{x}^l$) as before are used when analysing clustering characteristics.
Due to the high dimensionality of the representations and the computational complexity associated with the high dimensionality, we reduce the representations to $\mathbf{X}^l \in \mathbb{R}^{N \times C}$.
This is achieved by taking the average of the elements in the feature map across the height $H$ and width $W$ for each channel, expressed as:
\begin{equation}
    \mathbf{X}^l_{n,c} = \frac{1}{H \cdot W} \sum_{h=1}^{H} \sum_{w=1}^{W} \mathbf{X}^l_{n, c, h, w}
\end{equation}
where $\mathbf{X}^l_{n, c, h, w}$ denotes the element of the tensor $\mathbf{X}^l$ at position $(n, c, h, w)$, and $\mathbf{X}^l_{n,c} \in \mathbb{R}^{N \times C}$ is the resulting 2D tensor. 
Each element represents the average value of the feature map over the spatial dimensions $H$ and $W$ for each channel $C$ and sample $N$.
We then normalize each sample representation by dividing it with its Euclidean norm to produce $\mathbf{x^l}$ to be $\in [0, 1]$. 

We consider cluster purity from both a class-based and class-agnostic approach. For the class-based approach, we do not apply any clustering techniques to the representations, but instead, we analyze the clusters naturally formed by the network for each labelled class. In this context, we measure the DBI of the layer representations using the labels of the sample classes as the cluster labels, following the method of Natekar and Sharma~\cite{natekar2020representation}.

For the class-agnostic approach, we first use the k-means clustering algorithm to group similar representations and provide cluster labels for each representation. To select the optimal number of clusters we first use 2 to 15 clusters during an initial clustering step. We then measure the purity of these representation clusters using the DBI metric, as discussed in Section \ref{sec:Clustering_Analysis_Class_Agnostic}.  We then select the optimal number of clusters as the clusters that produced the lowest DBI score. 

\section{Sparsity Analysis and Results}
\label{sec:Analysis_sparsity}

In this section, we analyze and compare the sparsity of networks trained with and without BatchNorm. We consider the sparsity of representations both per channel (Section~\ref{sec:Analysis_sparsity_channel}) and per layer (Section~\ref{sec:Analysis_sparsity_layer}). 

\subsection{Channel Sparsity}
\label{sec:Analysis_sparsity_channel}

We begin by analyzing the sparsity of individual channels within the network for selected layers. 
The analysis uses the representations of the hidden layers with dimensions $\mathbf{x}^l \in \mathbb{R}^{C \times NHW}$, as described in Section \ref{sec:Exp_sparsity}. The sparsity for each layer is calculated according to Equation \ref{eq:Sparsity_avg_channel}, using $5\ 000$ randomly selected training samples. 
We select a subset of layers spread through the networks and evaluate the same layers per model type.

\begin{figure}[th]
    \centering
    \begin{subfigure}[t]{0.3\textwidth}
        \centering
        \includegraphics[width=\linewidth]{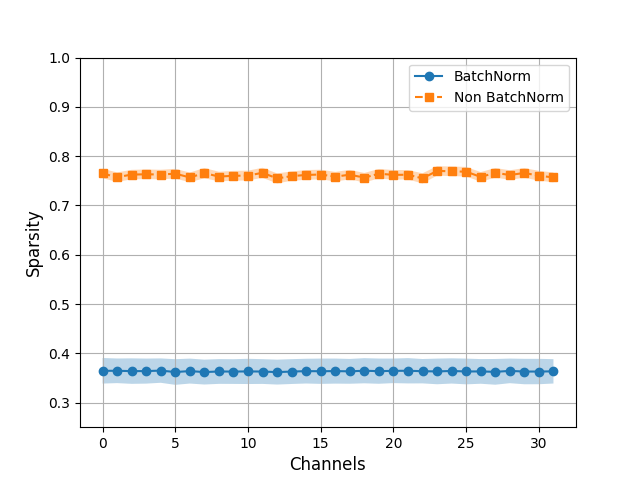}
        \caption{Layer $2$}
    \end{subfigure}
    \begin{subfigure}[t]{0.3\textwidth}
        \centering
        \includegraphics[width=\linewidth]{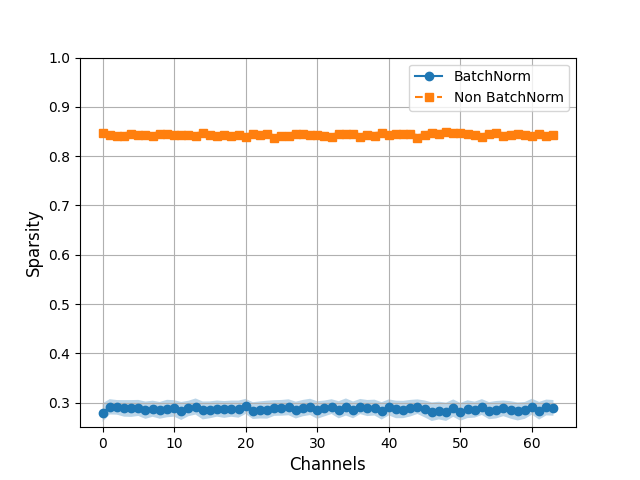}
        \caption{Layer $3$}
    \end{subfigure}
    \begin{subfigure}[t]{0.3\textwidth}
        \centering
        \includegraphics[width=\linewidth]{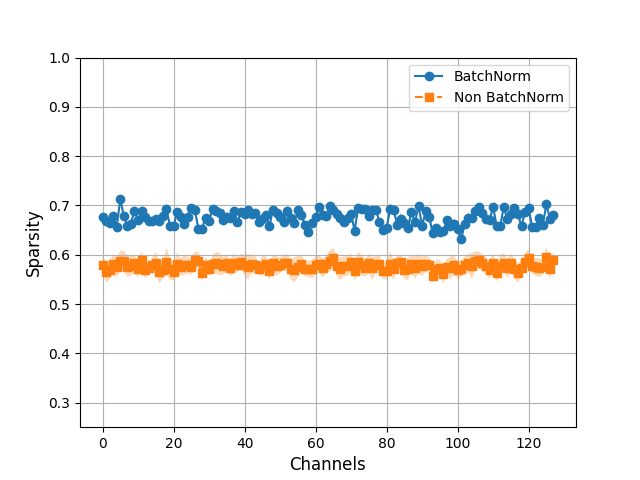}
        \caption{Layer $4$}
    \end{subfigure} 
    \caption{Average channel sparsity for BatchNorm (blue) and Non-BatchNorm (orange) models for selected convolutional layers of the standard CNN models trained on MNIST. Error bars show the standard deviation across $4$ random initialization seeds.}
    \label{fig:MNIST_Channel_Sparsity}
\end{figure}

\begin{figure}[th]
    \centering
    \begin{subfigure}[t]{0.3\textwidth}
        \centering
        \includegraphics[width=\linewidth]{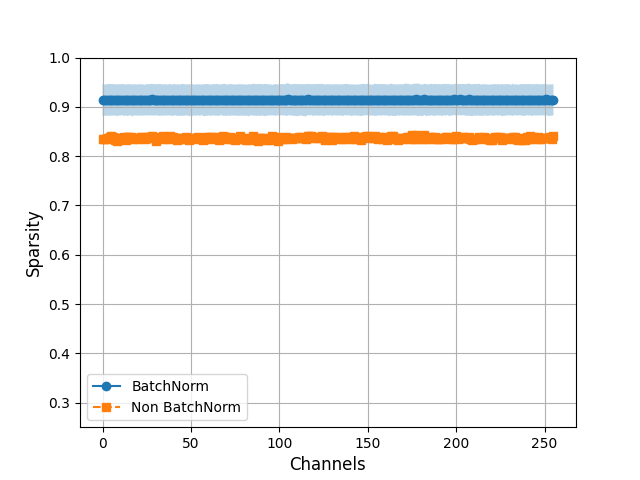}
        \caption{Layer $6$}
    \end{subfigure}
    \begin{subfigure}[t]{0.3\textwidth}
        \centering
        \includegraphics[width=\linewidth]{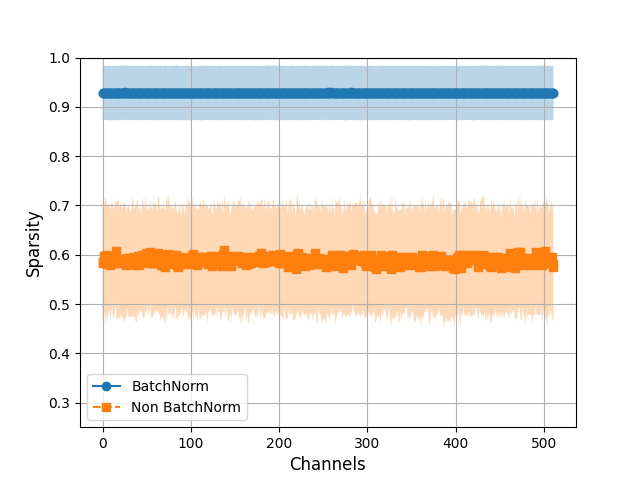}
        \caption{Layer $10$}
    \end{subfigure}
    \begin{subfigure}[t]{0.3\textwidth}
        \centering
        \includegraphics[width=\linewidth]{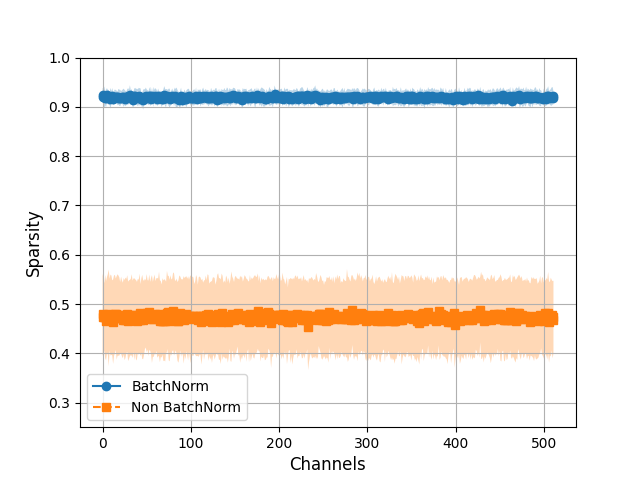}
        \caption{Layer $13$}
    \end{subfigure}
    \caption{Average channel sparsity for BatchNorm (blue) and Non-BatchNorm (orange) models in selected VGG-16 convolutional layers trained on CIFAR10, with error bars showing the standard deviation across $4$ random initialization seeds.}
    \label{fig:CIFAR_Channel_Sparsity}
\end{figure}

Figures \ref{fig:MNIST_Channel_Sparsity} and \ref{fig:CIFAR_Channel_Sparsity} illustrate the average channel sparsity for the four models for both BatchNorm and Non-BatchNorm configurations, across the MNIST and CIFAR10 datasets, respectively. The standard deviation of metrics across different seeds is indicated with shading. We choose layers $6, 10$ and $13$ to probe the networks in the deeper layers at consistent intervals.

%
For the standard CNN models trained on MNIST, as shown in Figure \ref{fig:MNIST_Channel_Sparsity}, we observe that the models trained with BatchNorm generally exhibit lower channel sparsity compared to those trained without BatchNorm. This observation is consistent across all channels and layers of the model, with the exception of the final layer, where channels in the BatchNorm models are more sparse than those in the Non-BatchNorm models.
We note that channel sparsity remains consistent across different initialization seeds, 
and are also relatively uniform within each layer. The one exception is the last layer, where there is a slightly larger degree of variation in channel sparsity, especially for BatchNorm models.


However, when we turn our attention to Figure~\ref{fig:CIFAR_Channel_Sparsity}, which shows the sparsity per channel for the VGG-16 models trained on CIFAR10, the opposite behaviour is observed. 
The VGG-16 models trained with BatchNorm tend to be more sparse than those trained without. 
We also note greater variation among different initialization seeds, especially for the Non-BatchNorm models. That said, we again note that each channel shows similar sparsity within a layer. 

When comparing Figures \ref{fig:MNIST_Channel_Sparsity} and \ref{fig:CIFAR_Channel_Sparsity}, we observe that the channels in the CIFAR10 models trained with BatchNorm are significantly more sparse than their MNIST counterparts. This could potentially be an artefact of the different datasets, as MNIST consists of many $0$-valued inputs in itself whereas CIFAR10 consists of more uniformly distributed values larger than $0$. It is therefore strange that the CIFAR10 BatchNorm models are \textit{more} sparse than those of MNIST.

\subsection{Layer Sparsity}
\label{sec:Analysis_sparsity_layer}

In this section, instead of analysing a single channel individually, we analyse the sparsity of each layer within the networks. 
This analysis uses the representations of the hidden layers with dimensions $\mathbf{x}^l \in \mathbb{R}^{N \times CHW}$, as described in Section \ref{sec:Exp_sparsity}. We once again make use of $5\ 000$ randomly selected training samples.
For each layer, the sparsity is computed according to Equation \ref{eq:Sparsity_avg_layer}.

\begin{figure}[th]
    \centering
    \begin{subfigure}[t]{0.48\textwidth}
        \centering
        \includegraphics[width=\linewidth]{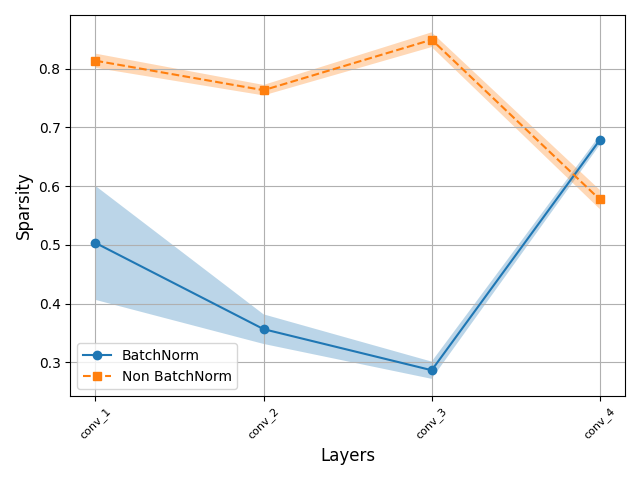}
        \caption{Standard CNN trained on MNIST}
    \end{subfigure}
    \begin{subfigure}[t]{0.48\textwidth}
        \centering
        \includegraphics[width=\linewidth]{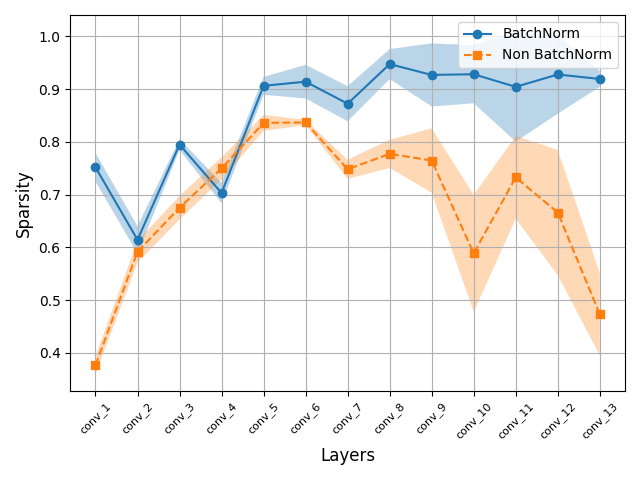}
        \caption{VGG-16 trained on CIFAR10}
    \end{subfigure}
    \caption{Average layer sparsity for models trained with (blue) and without (orange) BatchNorm. Error bars indicate the standard deviation across $4$ random initialization seeds.}
    \label{fig:Layer_Sparsity}
\end{figure}

Figure \ref{fig:Layer_Sparsity} illustrates the average layer sparsity for both the MNIST (left) and CIFAR10 (right) architectures with and without BatchNorm. We again include error bars indicating the standard deviation across the $4$ random initialization seeds.
%
%
Figure \ref{fig:Layer_Sparsity} reveals a pronounced difference in layer sparsity between the standard CNNs and VGG-16 models. 
For the standard CNNs, the representations of the BatchNorm models are generally less sparse than those of the models trained without BatchNorm, with the exception of the final layer. 
Conversely, for the VGG-16 models, the BatchNorm models exhibit greater overall sparsity compared to the Non-BatchNorm models.
The sparsity values are consistent across different seeds for both architectures, except for the last layers of the VGG-16 models where the variation is higher between seeds. 
Notably, the sparsity in the VGG-16 Non-BatchNorm models decreases significantly in the final layers, while the sparsity in the BatchNorm models remains consistent and relatively high.

It has been speculated that sparser representations are more desirable, as they suggest that the network is selectively activating in response to specific features. However, it is evident that BatchNorm neither consistently induces nor prevents sparsity in the representations.
Yet, in both cases, the models with BatchNorm generalize better. This points to the notion that the presence or absence of sparse representations does not appear to be a primary factor that determines the generalization ability of a network. 
Instead, it is likely that other effects of BatchNorm play a more significant role in the network's generalization performance.



\section{Clustering Analysis and Results}
\label{sec:Clustering_Analysis}
In this section, we analyse the effect of BatchNorm on the purity of representational clusters formed by the network. We consider this from both a class-specific (Section~\ref{sec:Clustering_Analysis_Class_Based}) and class-agnostic (Section~\ref{sec:Clustering_Analysis_Class_Agnostic}) perspective.

\subsection{Class-based Analysis}
\label{sec:Clustering_Analysis_Class_Based}

\begin{figure}[tbh]
    \centering
    \begin{subfigure}[t]{0.48\textwidth}
        \centering
        \includegraphics[width=\linewidth]{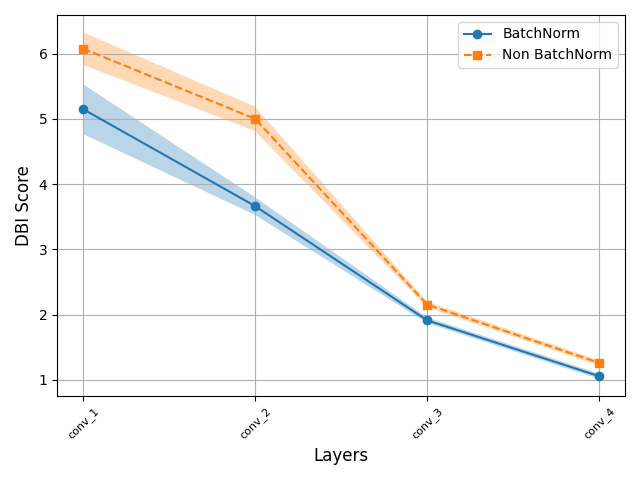}
        \caption{Standard CNN trained on MNIST}
    \end{subfigure}
    \begin{subfigure}[t]{0.48\textwidth}
        \centering
        \includegraphics[width=\linewidth]{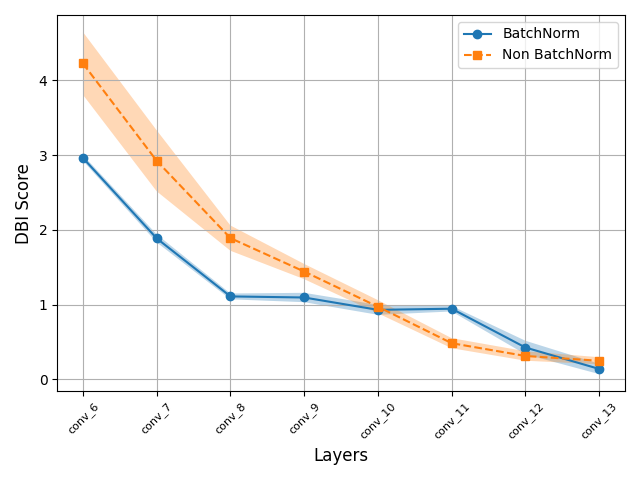}
        \caption{VGG-16 trained on CIFAR10}
    \end{subfigure}
    \caption{Average class-based DBI cluster purity score per layer for BatchNorm (blue) and Non-BatchNorm (orange) models, with error bars showing standard deviation across $4$ random initialization seeds. Lower scores indicate higher purity.}
    \label{fig:Activation_DBI}
\end{figure}

We first analyze the purity of the clusters formed in each layer of the network, under the assumption that the class of each sample corresponds to the cluster to which it belongs. 
We use $15\ 000$ randomly selected training samples and preprocess the representations as described in Section~\ref{sec:Exp_cluster}. 
%
Figure \ref{fig:Activation_DBI} presents the DBI scores for each layer of the networks trained with and without BatchNorm, for both the MNIST and CIFAR10 architectures, respectively.
The average DBI scores across the $4$ different initialization seeds, along with the corresponding standard deviation, are displayed.
For the VGG-16 network, we only show from convolutional layer six onward as the earlier layers are not informative and show unstable behavior among different seeds.

For both network architectures, the models trained with BatchNorm generally exhibit higher cluster purity compared to those trained without. 
An exception is observed in layers $11$ and $12$ of the VGG-16 model; however, the overall cluster purity remains higher in the BatchNorm models. Furthermore, we also observe that the purity values are relatively consistent across the different seeds, with minimal variation.
Since only the $10$ dataset classes serve as the cluster labels, and each sample can only belong to the cluster corresponding to its class label, both models start off with very low purity when traversing through the layers. However, in the deeper layers, clusters become increasingly pure, ultimately nearing complete purity, as observed in layer $13$ of the VGG-16 models.

We suspect that the common features shared among different classes could lead to the formation of either fewer or more clusters than there are classes in the earlier layers. Overall, observations are consistent with the well-known belief that earlier layers learn more general patterns in CNNs~\cite{caron2018deep}.
In deeper layers, we suspect that more class-specific features, with fewer shared features between classes, likely contribute to the higher clustering purity observed. 

\subsection{Class-agnostic Analysis}
\label{sec:Clustering_Analysis_Class_Agnostic}

We now turn our attention to the class-agnostic clustering perspective. 
As described in Section \ref{sec:Exp_cluster}, we assign clusters using the k-means clustering algorithm, and then measure the resulting purity. 
As with the previous experiment, a total of $15\ 000$ randomly selected training samples are utilized for these calculations.

\subsubsection{Optimal number of clusters}

\begin{figure}[th]
    \centering
        \begin{subfigure}[t]{0.48\textwidth}
        \centering
        \includegraphics[width=\linewidth]{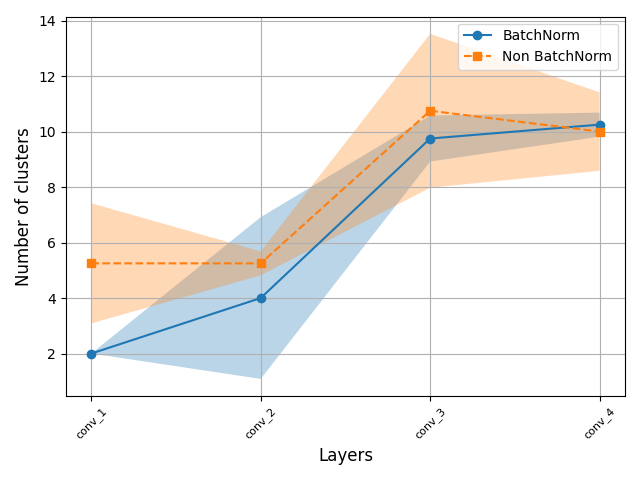}
        \caption{MNIST trained standard CNN}
    \end{subfigure}
    \begin{subfigure}[t]{0.48\textwidth}
        \centering
        \includegraphics[width=\linewidth]{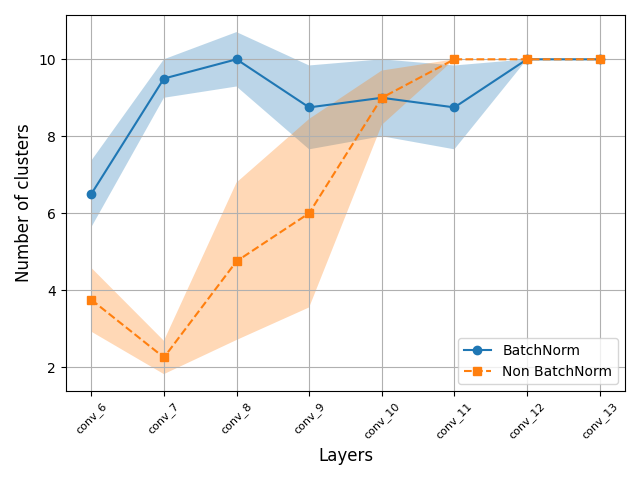}
        \caption{CIFAR10 trained VGG-16}
    \end{subfigure}
    \caption{Average optimal number of class-agnostic clusters per layer for BatchNorm (blue) and Non-BatchNorm (orange) models, with error bars showing the standard deviation across $4$ random initialization seeds.}
    \label{fig:Num_cluster_labels}
\end{figure}

The optimal number of clusters for each layer is shown in Figure~\ref{fig:Num_cluster_labels} for both the MNIST (left) and CIFAR10 (right) architectures. The error bars again report the standard deviation across the $4$ random initialization seeds.
We only show from convolutional layer $6$ onward for the VGG-16 models as the initial layers are not informative and unstable between seeds.
%
In Figure~\ref{fig:Num_cluster_labels} we observe that, for both architectures, the optimal number of clusters initially starts small but then increases. At the final layers, the optimal number converges to $10$ clusters, corresponding to the number of classes in the dataset.
For the standard CNN models, the increase in the optimal number of clusters occurs in the middle of the network; for the VGG-16 models, this occurs closer to the end. 
This is true for both the models trained with and without BatchNorm.

For VGG-16 models, the optimal cluster patterns differ across model types
In BatchNorm models, the number of optimal clusters increases in the early layers, stabilizing until the last two layers, where it matches the number of classes.
For Non-BatchNorm models, the optimal number of clusters is initially reached but significantly decreases in the middle layers, then increases again in the later layers, ultimately matching the number of classes in the final layers.

These observations support the hypothesis presented in Section \ref{sec:Clustering_Analysis_Class_Based} that optimal class-wise clusters are achieved only in the later layers. It suggests that while different class features contribute to cluster formation, the clusters do not become class-specific until the final layers of the network.
These plots also suggest that the improved generalization ability of BatchNorm models is likely due to the greater consistency of clusters formed by the representations throughout the network, compared to those in models trained without BatchNorm.

\subsubsection{Cluster purity}

\begin{figure}[tbh]
    \centering
        \centering
        \begin{subfigure}[t]{0.48\textwidth}
        \centering
        \includegraphics[width=\linewidth]{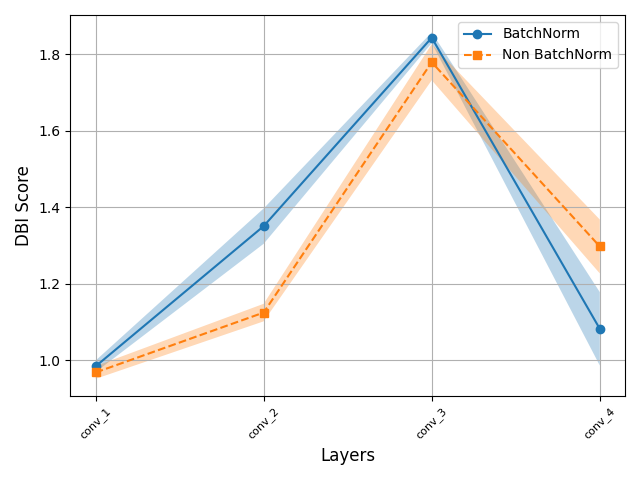}
        \caption{MNIST trained standard CNN}
    \end{subfigure}
    \begin{subfigure}[t]{0.48\textwidth}
        \centering
        \includegraphics[width=\linewidth]{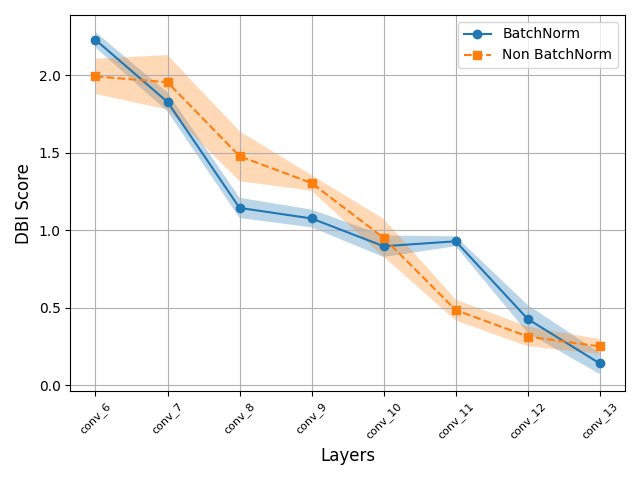}
        \caption{CIFAR10 trained VGG-16}
    \end{subfigure}
    \caption{Average class-agnostic DBI cluster purity score for k-means per layer for BatchNorm (blue) and Non-BatchNorm (orange) models, with error bars indicating standard deviation across $4$ random initialization seeds. Lower scores indicate higher purity.}
    \label{fig:DBI_cluster_k_means}
\end{figure}

We now consider the DBI score of each optimal cluster found using k-means. This is shown for the MNIST and CIFAR10 models in Figure~\ref{fig:DBI_cluster_k_means}.
We again only show from convolutional layer $6$ onward for the VGG-16 models as the initial layers are not informative and unstable between seeds.
In Figure \ref{fig:DBI_cluster_k_means} we observe that the standard CNNs, for both model types, follow the same pattern where the DBI score increases significantly until layer $3$ and then decreases. 
The BatchNorm models also have a slightly higher score than the Non-BatchNorm except for the last layer where the BatchNorm has a lower score.
Comparing these results with what was seen in Figure \ref{fig:Activation_DBI} with Figure \ref{fig:DBI_cluster_k_means} there is a difference: the scores in Figure \ref{fig:Activation_DBI} only decreases whereas the DBI scores in Figure \ref{fig:DBI_cluster_k_means} increases before decreasing in the last layer.
One thing to note is that the DBI scores in Figure \ref{fig:Activation_DBI} are significantly higher (poorer) than those in Figure \ref{fig:DBI_cluster_k_means}.


For the VGG-16 network, the overall DBI score of the BatchNorm models is better compared to the non-BatchNorm models. Variation between seeds is more stable for the BatchNorm models than non-BatchNorm models. This suggests that the clusters are more stable for the BatchNorm models over seeds.
Comparing Figure \ref{fig:DBI_cluster_k_means} with Figure \ref{fig:Activation_DBI}, the same pattern emerges for the VGG-16 models where the BatchNorm models have purer clusters except for layers 11 and 12. This suggests better cluster forming in general, and thus better generalization in the BatchNorm models, especially in the last layer.


\section{Discussion}

Let us summarise the main observations. In the context of convolutional neural networks and the experiments conducted here:
\begin{enumerate}
    \item BatchNorm does not consistently induce nor prevent sparsity in hidden representations.
    \item Sparse representations do not correlate with better generalization.
    \item BatchNorm tends to produce purer representational clusters than models trained without, especially in the case of class-based representations. For class-agnostic representation, results are less clear, except at the final layer, where BatchNorm clusters are again purer.
    \item BatchNorm forms class-agnostic clusters earlier in the network and these early clusters are more consistent with the actual number of classes.
\end{enumerate}

It is challenging to determine the exact cause of the sparsity in the representations studied because we observed significant differences between models trained on MNIST and those trained on CIFAR10. However, a clear distinction exists between models with BatchNorm and those without.
For standard CNNs, applying BatchNorm results in much less sparse representations, which aligns with the findings of Pretorius et al.~\cite{pretorius2019relu}. 
In contrast, VGG-16 models exhibit the opposite behaviour, where BatchNorm models produce significantly more sparse representations than their non-BatchNorm counterparts.


The difference in sparsity might be influenced by the dataset on which the models are trained, or to the architectural differences between the models. Specifically, VGG-16 models with BatchNorm seem to activate for more specific features, leading to increased sparsity, particularly around layer 5. On the other hand, non-BatchNorm models in VGG-16 also show increased sparsity until layer $5$ but revert to activating more general features in deeper layers, causing a reduction in sparsity.
%
However, note the variability in the results across seeds for the deeper layers -- this could be indicative of other underlying processes at play.

Generally, BatchNorm models tend to produce purer clusters compared to non-BatchNorm models. In the VGG-16 models, both clustering methods yielded similar results with BatchNorm performing better overall, with the exception of 
layers $11$ and $12$.
For the layers where non-BatchNorm perform better in class-agnostic clustering, it results in more clusters and a wider range of cluster numbers across seeds compared to the BatchNorm models.


The work also raises additional questions: 
Do clearer patterns emerge if the experiment is repeated on additional datasets and architectures? 
Since the representations are highly dimensional, clustering them is computationally expensive, and repeating these experiments over multiple seeds quickly becomes expensive, it remains an avenue we wish to pursue further.
Similarly, we used $0$ as the threshold for measuring sparsity, even though `very small' numbers could also be considered sparse. In the current analysis, we avoided re-running experiments across multiple thresholds, although it could be worth exploring in more detail.

\section{Conclusion} 

Studying the characteristics of DNNs' internal representations can provide insight into the underlying mechanisms that contribute to good generalization.
In this study, representations were analysed from two convolutional architectures: a small fairly standard CNN architecture and a VGG-16 architecture. 
We developed models with similar training accuracy but -- since the models with BatchNorm generalized better on unseen data -- the models represented different generalization gaps.
The main question asked was: how does BatchNorm affect the characteristics of the internal representations of these two sets of models? 

 Representational sparsity did not correlate either with the generalization ability of the networks or could be linked to whether BatchNorm was applied or not. 
 Clearer trends emerged with regard to the consistency of representations, specifically when considering the purity of clusters formed by these representations at different layers in the network.
Overall, BatchNorm tended to produce purer representational clusters, both in the case of class-based and class-agnostic analysis.
In addition, class-specific clusters formed earlier in the BatchNorm networks, and these were more consistent with the actual number of classes.


Trends observed were complex and pose additional questions that can be explored further.
In this work, our objective was to provide initial results and motivate the importance and potential of studying the characteristics of DNN representations to shed light on the underlying mechanisms controlling the generalization ability of these models. 

\subsubsection*{Acknowledgements}

This work is based on the research supported in part by the National Research Foundation of South Africa (Ref Numbers PSTD23042296065, RA211019646111).
\newpage

\bibliographystyle{unsrt}  
\bibliography{references}  






\end{document}